\documentclass[letterpaper, 10 pt, conference]{ieeeconf}  

\usepackage[space]{cite}
\usepackage[utf8]{inputenc}
\usepackage[T1]{fontenc}
\usepackage{times}
\usepackage{epsfig}
\usepackage{graphicx}
\usepackage{amsmath}
\usepackage{amssymb}
\usepackage{graphics} 
\usepackage{mathtools}
\usepackage{subcaption}
\usepackage{booktabs}
\usepackage{adjustbox}
\usepackage{caption}
\usepackage{stfloats}
\usepackage{multirow}
\usepackage{color}
\usepackage[linesnumbered,ruled,lined,commentsnumbered]{algorithm2e}
\usepackage[pagebackref=false,breaklinks=true,colorlinks,bookmarks=false]{hyperref}
\usepackage{algpseudocode}
\usepackage{tabularray}

\newcommand{\note}[1]{{\color{red}{#1}}}

\IEEEoverridecommandlockouts                              

\title{\LARGE \bf Redundant and Loosely Coupled LiDAR-Wi-Fi Integration for Robust Global Localization in Autonomous Mobile Robotics}

\author{Nikolaos Stathoulopoulos$^{1}$, Emanuele Pagliari$^{2}$, Luca Davoli$^{2}$, George Nikolakopoulos$^{1}$
\thanks{This work has been funded by the European Unions Horizon 2020 Research and Innovation Programme under Grant Agreement No. 101003591, NEX-GEN SIMS, and Grant Agreement No. 876019, ADACORSA.}
\thanks{$^{1}$The Authors are with the Robotics and AI Group, Department of Computer, Electrical and Space Engineering, Lule\r{a} University of Technology, 97187 Lule\r{a}, Sweden.} %
\thanks{$^{2}$The Authors are with the Internet of Things (IoT) Lab, Department of Engineering and Architecture, University of Parma, 43124 Parma, Italy.}
\thanks{Corresponding Author's Email: \texttt{niksta@ltu.se}}}

\begin{document}

\maketitle
\thispagestyle{empty}
\pagestyle{empty}

\begin{abstract}
This paper presents a framework addressing the challenge of global localization in autonomous mobile robotics by integrating LiDAR-based descriptors and Wi-Fi fingerprinting in a pre-mapped environment. This is motivated by the increasing demand for reliable localization in complex scenarios, such as urban areas or underground mines, requiring robust systems able to overcome limitations faced by traditional Global Navigation Satellite System (GNSS)-based localization methods. By leveraging the complementary strengths of LiDAR and Wi-Fi sensors used to generate predictions and evaluate the confidence of each prediction as an indicator of potential degradation, we propose a redundancy-based approach that enhances the system's overall robustness and accuracy. The proposed framework allows independent operation of the LiDAR and Wi-Fi sensors, ensuring system redundancy. By combining the predictions while considering their confidence levels, we achieve enhanced and consistent performance in localization tasks. 
\end{abstract}


\section{INTRODUCTION} \label{sec:intro}
Global localization plays a crucial role in autonomous mobile robotics from several perspectives: providing the necessary foundation for localization algorithms; enabling robots to re-establish their positions after leaving a mapped area; restarting missions in previously mapped environments; mitigating pose estimation drift through loop-closure detection~\cite{loop_closure}; and merging mapping data collected during different sessions~\cite{stathoulopoulosFRAME2023}. However, certain environments, including urban areas, industrial settings and underground mines, present unique challenges for traditional GNSS-based localization systems, often arising from issues like multi-path effects and limited satellite reception. 
The ability to operate within a global map is essential for successful exploration and navigation missions. A global map provides robots with valuable information for various tasks, such as path planning, coordination of multiple robots, and localization of objects and survivors during Search And Rescue (SAR) missions. However, in these complex scenarios, localization algorithms can face temporary failures. Factors such as sensor faults, presence of dust particles or even drifting, can cause these algorithms to temporarily lose accuracy. Consequently, the robot's current pose within the global map becomes misaligned, thus potentially affecting its ability to effectively carry out its mission objectives. 

With the recent advancements in efficient data representations and feature descriptors for LiDAR point clouds through Deep Learning (DL) techniques, LiDAR sensors have shown promising results in the realm of computer vision (especially in the aim of place recognition). While LiDAR-based systems offer advantages in terms of immunity to appearance changes and illumination, these sensors can face difficulties in accurately capturing certain types of surfaces, such as transparent or reflective objects. Additionally, adverse weather conditions (like rain, fog, or dust) can affect their performance
\begin{figure}[t!]
    \centering
    \includegraphics[width=\columnwidth]{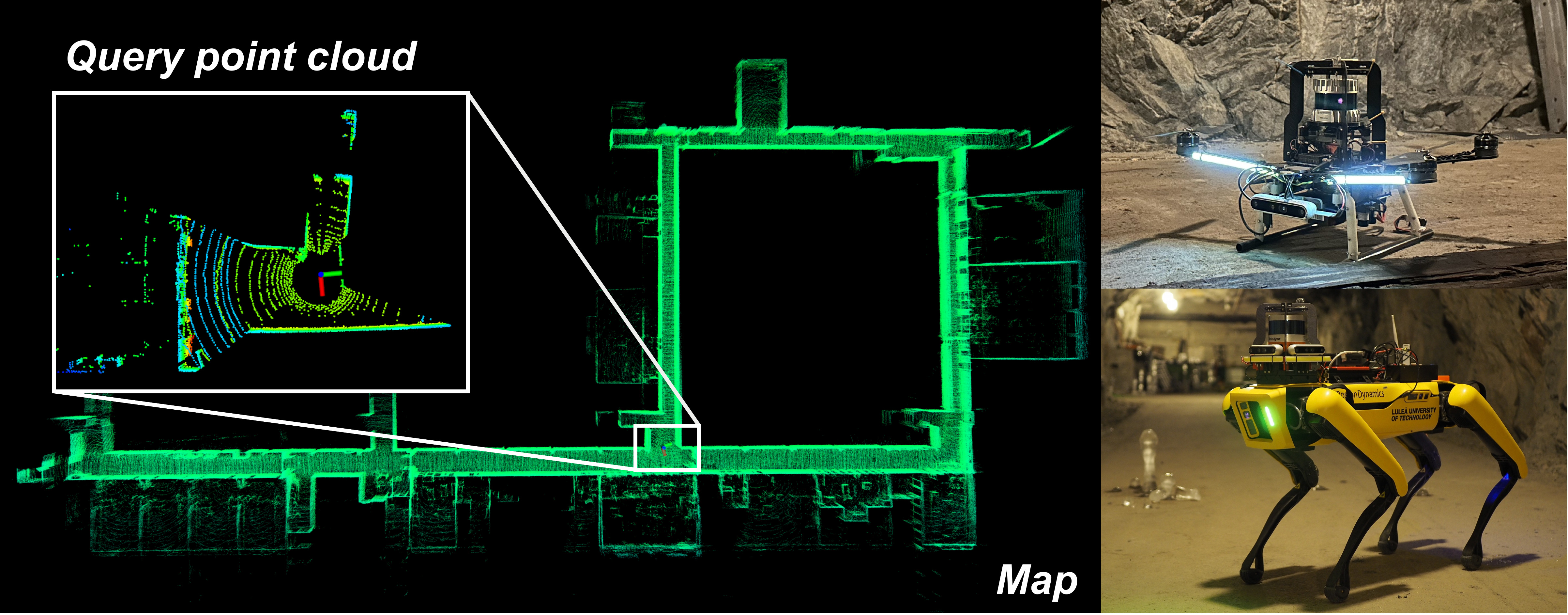}
    \caption{Point cloud map generated from an experiment showcased in this article, demonstrating the queried point cloud, and showcasing several mobile robots equipped with the novel and lightweight proposed solution.}
    \label{fig:introduction}
    \vspace{-4mm}
\end{figure}
Hence, in order to address these challenges and enhance the reliability of localization systems, redundancy becomes crucial. By incorporating additional sensors, such as Wi-Fi sensors, alongside LiDAR, the system can benefit from complementary information sources. It is well-know how, in the last decade, the use of the IEEE~$802.11$~Wi-Fi technology has further spread, enriching the overall availability and coverage of wireless networks, thus making it more feasible and easier to exploit existing networks for different applications. Wi-Fi technology can provide valuable data for localization, especially in scenarios where LiDAR may face limitations or obstacles. This redundancy increases the overall robustness and reliability of the system, enabling more comprehensive and accurate localization capabilities. 

Given the above remarks, in this paper we present a framework integrating LiDAR-based descriptors and Wi-Fi fingerprinting to address the challenge of global localization in a pre-mapped environment. In particular, we emphasize the system's redundancy by maintaining the independence of these two sensors, but at the same time leveraging them to generate predictions, and then we evaluate the confidence of each prediction as an indicator of potential degradation. By merging the two predictions while considering their confidence levels, we demonstrate an enhanced performance. 

The contributions of this paper can be summarized as follows: 
(i) the development of a compact and affordable Wi-Fi fingerprinting solution specifically designed for mobile robots, enabling accurate global localization capabilities;
(ii) the establishment of a redundancy-based system architecture that enhances the reliability and robustness of the overall framework, leveraging low-cost resources to improve LiDAR-based descriptors; 
(iii) significantly improved global localization likelihood in challenging environments, where traditional methods tend to underperform due to perceptual degradation, through the integration of LiDAR-based descriptors and Wi-Fi fingerprinting; 
(iv) promising results demonstrating the real-world applicability of the proposed framework in inspection and monitoring scenarios, showcasing its potential for practical implementation.


\section{RELATED WORK}
\label{sec:relatedworks}


In this section, we review the relevant literature in two key areas: (i) LiDAR-based place recognition and (ii) Wi-Fi fingerprinting for localization. 

In the field of computer vision and place recognition, camera-based approaches have long dominated the scene~\cite{cadena2010}. One of the widely known descriptors is SURF~\cite{Bay_2006}, which utilizes visual features and geometric transformations to establish correspondences between images, enabling accurate localization. However, the limitations of cameras in low light conditions, weather changes, and the lack of depth information have led to an increasing interest in LiDAR-based methods. LiDAR sensors gained appeal due to their immunity to lighting variations and their ability to provide rich 3D information about the environment, and became particularly valuable for applications where accurate localization is crucial, such as autonomous driving or mapping. In~\cite{pointnetVLAD2018}, a framework denoted as PointNetVLAD, which first processes each point of a point cloud individually using PointNet to extract local features, and then aggregates these local features using VLAD, which encodes a global representation of the entire point cloud, has been proposed. In detail, VLAD represents each cluster of local features by computing the residual vectors between each local feature and a set of learned cluster centers. The authors of OREOS~\cite{oreosSchaupp2019} proposed a novel approach to address the heavy computational load of 3D LiDAR scans, by projecting them onto a 2D plane while preserving depth information. This technique reduces the computational burden, making LiDAR more suitable for mobile robots or UAVs without sacrificing the advantages of LiDAR-based place recognition. In OverlapNet~\cite{chen2021auro}, a step further has been performed by exploiting different types of information generated from LiDAR scans to provide overlap and relative yaw angle estimates between pairs of 3D scans. The range images are enhanced with information such as normals, intensity and semantic data.

The approach of Wi-Fi fingerprinting for indoor localization has been adopted for several applications, spacing from users' tracking through their smartphones~\cite{mobile2016}, up to the localization of IoT devices in industrial environments~\cite{iotfinger2016}. This solution is well known for its low implementation cost, especially on new devices (e.g., smartphones, IoT devices, etc.) which nowadays all integrate Wi-Fi connectivity among other communication protocols. An overview of existing Wi-Fi fingerprinting solutions for localization has been carried out in~\cite{IEToverview2022, ieeeMLwifi2021}, where several localization algorithms using Wi-Fi fingerprints have been illustrated, showing how the use of different techniques and Machine Learning (ML) algorithms can achieve an accuracy of a few meters for indoor localization applications. 

However, for autonomous robotics applications, a step forward is needed in order to allow the execution of complex missions in harsh environments. A possible solution can be the fusion of heterogeneous data sources for localization purposes, such as, as an example, LiDAR, IMU and visual odometry algorithms, to be fused with Wi-Fi fingerprints. In~\cite{sensors2022lrfwivi}, Wi-Fi fingerprinting---although focusing on the Channel State Information (CSI) instead of the Received Signal Strength Index (RSSI)---has been combined with visual SLAM algorithm. Despite the promising results and relative low cost of the platform, the proposed solution needs a specific Wi-Fi network setup to collect the CSI fingerprints, since this metric is not supported by all the Wi-Fi Access Points (APs), therefore the solution cannot exploit the Wi-Fi networks available in the environment. 

Another similar method to fuse Wi-Fi fingerprinting---this time relying on RSSI and MAC addresses---together with visual SLAM algorithm, is proposed in~\cite{autonomousrobots2019}, highlighting the improvement that the combination of both technologies allows for indoor localization applications. However, the visual SLAM system is known to poorly work in low light conditions with reflections or dust, also showing a weak accuracy for environments with repetitive patterns (e.g., same wall geometry). Finally, in~\cite{case2022} an approach for a Wi-Fi fingerprinting and LiDAR SLAM fusion technique (similar to the one presented in our paper) is proposed, achieving interesting results, although the Wi-Fi data collection equipment detailed in~\cite{case2022} relies on the use of several smartphones as Wi-Fi scanners, which is impractical for an implementation on a constrained platform, such as a UAV. Also, during data collection, the used platform has been moving at a very low horizontal speed of~$0.4$~m/s, that, in favor of a richer and more complete Wi-Fi fingerprinting database, makes the initial data collection extremely time demanding, especially over a large area.


\section{PROBLEM FORMULATION}
\label{sec:problem}

When dealing with autonomous robotic applications, one of the most tedious problems is how to localize, in an indoor GNSS-denied environment, the position of the robotic platform with respect to an already mapped environment. While SLAM algorithms applied to the point cloud collected by the LiDAR can already achieve a good level of accuracy in many environments, they are still affected by the loop-closure issue and present poor performance in environments where dust and repetitive patterns in the surrounding are present, such as, for example, in a mining tunnel with walls made of rocks or in corridors with recurrent windows, doors or other fixed elements. 
These problems can heavily affect autonomous mobile robotic missions, where the platform has to localize itself in a global map previously built aiming to complete the assigned task. In fact, a failure of the localization estimation in the environment may lead to a failure in the mission, thus requiring an external human intervention.
Since Wi-Fi networks are nowadays deployed in almost every working environment, even in the most modern underground mine tunnels, a possible way to significantly reduce the risk of localization-related issues in autonomous missions might be further mitigated by enhancing LiDAR-based algorithms with the easy to integrate Wi-Fi fingerprinting techniques, thus allowing to integrate the point cloud maps with Radio Frequency (RF) data as a \textit{fourth dimension}.

\begin{figure*}[b!]
    \centering
    \includegraphics[width=\textwidth]{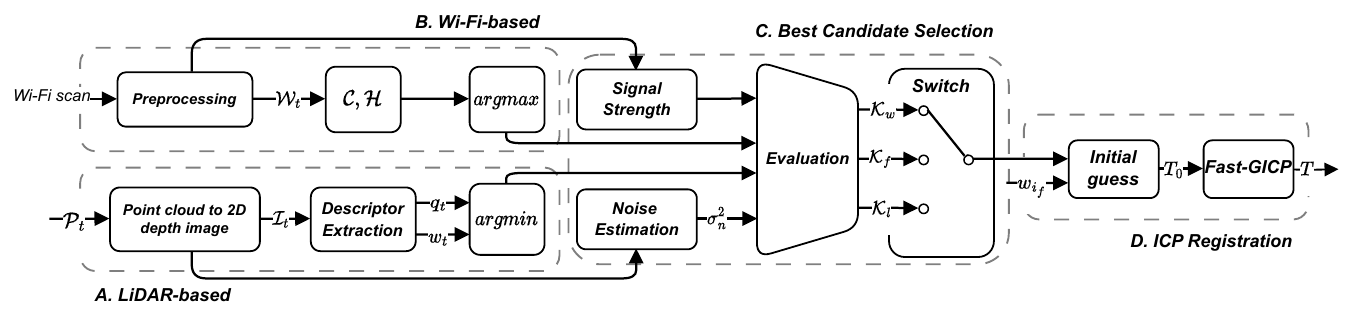}
    \caption{An overview of the proposed system architecture, divided into four main components. (A) LiDAR-based place recognition; (B) Wi-Fi fingerprinting; (C) best candidate selection; and (D) ICP Point Cloud Registration.}
    \label{fig:architecture}
\end{figure*}

The goal of this paper is to introduce a global localization algorithm able to yield a rigid transform $T \in SE(3)$ so that the current robot frame $\mathcal{R}$ is transformed to the global map frame $\mathcal{M}$. Given a map $M = \{ m_n |n = 1, 2, \ldots, N \}$ that is a set of 3D points $m_n \in \mathbb{R}^3$ and a trajectory $Tr = \{ p_k |k = 1, 2, \ldots, K \}$ that is a set of 3D poses $p_k = (x_k, y_k, z_k, \theta_k)$, we aim at extracting discriminative features from the observation sets $\mathcal{P}$ and $\mathcal{W}$. In detail, the set $\mathcal{P}$ contains the 3D LiDAR point cloud scans $\mathcal{P}_k = \{s_{k,l} | l = 1, 2, \ldots, L\}$, where $s_{k,l} \in \mathbb{R}^3$, while the set $\mathcal{W}$ contains the Wi-Fi attributes $\mathcal{W}_k = \{w_{k,l} | l = 1, 2, \ldots, L\}$. The global localization process can be expressed as a function $f$ of the current observations $\mathcal{P}_t$ and $\mathcal{W}_t$ as follows:
\begin{equation}
    x,y,z,\theta = f(\mathcal{P}_t, \mathcal{W}_t)\,.
\end{equation}
The homogeneous rigid transform can be constructed as:
\begin{equation} \label{eq:transform}
    T = \left[ \begin{array}{cc}
         R_z(\theta) & p \\
         0 & 1
    \end{array} \right] \in SE(3),
\end{equation}
where: $R_z(\theta) \in SO(3)$ is the rotational matrix of the yaw angle, and $p = (x,y,z)^\mathsf{T}$ is the translational vector. In order to address the aforementioned problem, we propose a combination of LiDAR-based place recognition loosely coupled with a Wi-Fi-based fingerprinting approach, shown in Fig.~\ref{fig:architecture} and detailed in Section~\ref{sec:methodology}.


\section{METHODOLOGY} \label{sec:methodology}
In order to solve the problem introduced in Section~\ref{sec:problem}, we divide the function $f$ into four components: (i) LiDAR-based place recognition; (ii) Wi-Fi fingerprinting; (iii) best candidate selection; and (iv) ICP Point Cloud Registration.
\subsection{LiDAR-based Place Recognition} \label{subsec:lidar}
With regard to the LiDAR-based place recognition, we decided to adopt an approach similar to the one proposed in~\cite{oreosSchaupp2019, stathoulopoulos20243deg}, since the transformation to the image space keeps the computational effort low. 
The first step is to transform the incoming LiDAR scans $\mathcal{P}_t$ in depth images $\mathcal{I}_t$ through a 2D spherical projection. 
Then, the images $\mathcal{I}_t$ are fed into the \textit{Descriptor Extraction} module, that outputs the $2 \times 64$ vectors $\vec q$ and $\vec w$. In particular, $\vec q$ is an orientation-invariant vector encoding place-dependent information and primarily used for querying similar point clouds, while $\vec w$ is an orientation-specific vector, responsible for regressing the yaw discrepancy between two point clouds. 
Then, in order to query the global map for near place candidates, we \textit{first} project the aforementioned LiDAR observations $\mathcal{P}$, yielding the range images $\mathcal{I}$. \textit{Then}, the \textit{Descriptor Extraction} module is responsible for generating the vector sets $Q$ and $W$, described by the function $D$ as $Q, W = D(\mathcal{I})$, where $Q$ and $W$ are the following vector sets:
\begin{align}
    Q=\, \{\vec q \in \mathbb{R}^{64}, \, k \in \mathbb{N}&: \vec q_1, \vec q_2, \dots, \vec q_k \} \\
    W=\, \{\vec w \in \mathbb{R}^{64}, \, k \in \mathbb{N}&: \vec w_1, \vec w_2, \dots, \vec w_k \}  \,.
\end{align}
In particular, the querying process can be described as a minimization problem where a \textit{k-d} tree is constructed with the vector set $Q$ and is searched through with the current vector $q_t$ to find the pair with the minimum distance in the $q$ vectors space:
\begin{equation} \label{eq:argmin}
    i = \operatorname*{arg\,min}_{i \, \in \, \mathbb{N}}f(Q,q_t) \,.
\end{equation}
As a common practice in the robotics and computer vision community, we keep the \textit{top-k} candidates, denoted as $\mathcal{K}_l = \{i_{l,1}, i_{l,2}, \ldots, i_{l,k}\}$, where the \textit{k} candidates correspond to the indexes of the most similar places and can be used to acquire the respective poses from the trajectory $Tr$.
\subsection{Wi-Fi Fingerprinting} \label{subsec:wifi_fingerprinting}
The next component is the Wi-Fi fingerprinting, whose integration details are discussed in Section~\ref{sec:experimental}. Given the observations $\mathcal{W}$ and the current Wi-Fi scan $\mathcal{W}_t$, we aim at finding the pair with the strongest correlation. Since a Wi-Fi scan contains multiple attributes, we currently use only two of them, in detail the MAC addresses of the scanned APs, denoted as $A = \{a_0, a_1, \ldots, a_N\}$, and the corresponding RSSI values, denoted as $S = \{s_0, s_1, \ldots, s_N\}$. Therefore, we can denote the Wi-Fi scan as $\mathcal{W}_t = \{A_t, S_t\}$. The amount of scanned APs is kept fixed at size $N$ at all times to make the calculations easier. Then: should the scanned APs be more than $N$, we would discard the ones with the highest RSSI; should the scanned APs be less than $N$, then we would pad them with a fixed value. The correlation between the observations $\mathcal{W}$ and the current scan $\mathcal{W}_t$ is given by the matrix $\mathcal{C}_{N \times N \times K}$, where $K$ is the total amount of observations in $\mathcal{W}$. In order to construct $\mathcal{C}$, we compare all the elements $A_k$ of $\mathcal{W}$ with $A_t$, and assign a value based on $S_k$ and $S_t$. This process can be described as follows: 
\begin{equation}
\mathcal{C}_{i,j}(\mathcal{W}_t, \mathcal{W}_k)=
    \begin{cases}
        \log_2 ( -S_{t,i} - S_{k,j}) &  A_{t,i} = A_{k,j} \\
        \hfil 0 &  A_{t,i} \neq A_{k,j} \,.
    \end{cases}
\end{equation}

\begin{figure}[b!]
    \centering
    \includegraphics[width=0.99\columnwidth]{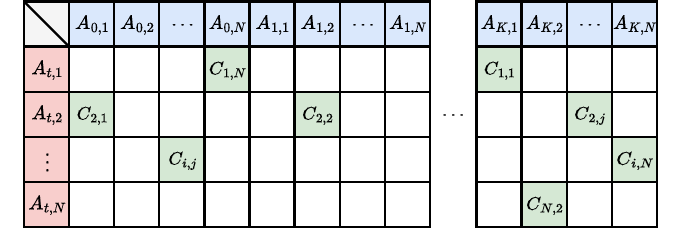}
    \caption{Example of correlation matrices $\mathcal{C}_{i,j}$, where the empty cells correspond to the case $A_{t,i} \neq A_{k,j}$, while the colored cells correspond to the case $A_{t,i} = A_{k,j}$.}
    \label{fig:correlation}
\end{figure}

An example of the correlation matrix is depicted in Fig.~\ref{fig:correlation}.
In the sequel, we seek to find the pair ($\mathcal{W}_k, \mathcal{W}_t$) with the highest correlation by summing up each matrix $\mathcal{C}_{k,i,j}$, yielding the vector $\mathcal{H}_{1 \times K}$, denoted as:
\begin{equation}
    \mathcal{H}_k = \sum_{i,j = 0}^{i,j = N} \mathcal{C}_{k,i,j}\;\;\mbox{for $k=0,1,\ldots,K$} \,.
\end{equation}
In the case of the Wi-Fi fingerprinting, the \textit{top-k} candidates are the ones with the maximum correlation sum, or simply:
\begin{equation}
    i = \operatorname*{arg\,max}_{i \, \in \, \mathbb{N}}\mathcal{H} \,.
\end{equation}\\
Similar to the LiDAR-based place recognition, we keep the \textit{top-k} candidates, denoted as $\mathcal{K}_w = \{i_{w,1}, i_{w,2}, \ldots, i_{w,k}\}$.
\subsection{Best Candidate Selection} \label{subsec:candidate_selection}
Before we select the best candidates out of the sets $\mathcal{K}_l$ and $\mathcal{K}_w$, we perform a quick evaluation of each candidate in order to add extra resiliency in the overall architecture. With regard to the LiDAR-based candidates, we estimate the presence of noise in the range images $\mathcal{I}_t,\mathcal{I}$ based on the technique proposed in~\cite{IMMERKAER1996300}. If the noise variance $\sigma_n^2 \geq \sigma^2_{threshold}$, then the candidates are considered a liability and are discarded. For the Wi-Fi-based candidates, we evaluate the correlation scores $C_{k,i,j}$. If the current RSSI values $S_t$ are considerably lower than the anchor values $S_k$, then the candidate should be disregarded due to low signal strength. As a final measure of assurance, we deploy a switching mechanism: in the case that the current LiDAR scan is too noisy, the candidate selection switches only to Wi-Fi; vice versa, if the Wi-Fi signal strength is too low, we switch only to the LiDAR candidates. After the candidate evaluation step, we make the assumption that the candidates with the highest likelihood $\mathcal{K}_f$ are the ones that exist in both sets, as follows:
\begin{equation}
    \mathcal{K}_f =
    \begin{cases}
        \hfil \mathcal{K}_w & \sigma_n^2 \geq \sigma^2_{threshold} \\
        \mathcal{K}_l \cap \mathcal{K}_w & \hfil otherwise \\
        \hfil \mathcal{K}_l & S_t \geq S_{threshold} \,.
    \end{cases}
\end{equation}
In reality, we expect the set $\mathcal{K}_w$ to constraint the predicted candidates of $\mathcal{K}_l$, since perceptual aliases in the LiDAR scans can lead to similar descriptors from similar but far away place candidates. 
It will be highlighted in Section~\ref{sec:experimental} how the final remaining candidates $\mathcal{K}_f$ from the two sets of \textit{top-10} candidates, are reduced to $1$-$3$ candidates.
\subsection{ICP Point Cloud Registration} \label{subsec:icp}
The final component of our proposed framework is the one responsible for yielding the complete homogeneous rigid transformation of the special Euclidean group $T \in SE(3)$. The best candidate $i_f$ corresponds to the pose $p_{i_f} \in Tr$. Initially, we use the orientation-specific vectors $\vec w_{i_f}$ and $\vec w_t$ in order to estimate the yaw discrepancy $\delta \theta$ between the current scan $\mathcal{P}_t$ and the queried scan $\mathcal{P}_{i_f}$ (see~\cite{oreosSchaupp2019}). Hence,  we can construct the initial translational vector as $(p_t$-$p_{i_f})^\mathsf{T}_0$ and the initial rotational matrix as $R_{z,0}(\delta \theta)$, used in the sequel as a prior to the Fast-GICP~\cite{fast_gicp} point cloud algorithm.


\section{INTEGRATION AND EXPERIMENTAL EVALUATION} \label{sec:experimental}
\begin{table*}[b!]
\caption{Recall score, mean distance error and standard deviation for the \textit{top-1} candidate, as seen on Fig. ~\ref{fig:recall_mean_std}.} \label{table:recall_mean_std}
\centering
\begin{tblr}{
  cells = {c},
  cell{1}{2} = {c=3}{0.239\linewidth},
  cell{1}{5} = {c=3}{0.239\linewidth},
  cell{1}{8} = {c=3}{0.239\linewidth},
  hline{1-3,6} = {-}{0.08em},
}
 & LTU - w/o added noise &  &  & LTU - with added noise &  &  & Underground mine &  & \\
 & RECALL (\%) & MEAN (m) & STD (m) & RECALL (\%) & MEAN (m) & STD (m) & RECALL (\%) & MEAN (m) & STD (m)\\
LiDAR-based & $88.7$  & $6.8$ & $16.2$ & $68.6$ & $16.4$ & $24.9$ & $86.6$ & $7.9$ & $25.1$ \\
Wi-Fi-based & $79.8$ & $4.2$ & $13.1$ & $79.8$ & $4.2$ & $13.1$ & $81.0$ & $3.8$ & $\textbf{4.7}$ \\
LiDAR+Wi-Fi & $\textbf{96.9}$ & $\textbf{3.3}$ & $\textbf{9.5}$ & $\textbf{91.4}$ & $\textbf{3.8}$ & $\textbf{10.2}$ & $\textbf{93.8}$ & $\textbf{3.4}$ & $11.6$
\end{tblr}
\end{table*}
The experimental evaluation of the proposed system can be divided into two phases: (i) an \textit{offline phase} and (ii) an \textit{online phase}. In the \textit{offline phase,} the environment is scanned with LiDAR and Wi-Fi scanners, gathering the point cloud of the environment together with the Wi-Fi data measurements. Then, these data are associated with the odometry computed by the robotic platform and, at the end of the survey, the final map of the environment is built. Once the data are gathered, the Wi-Fi fingerprints database is generated, associating each Wi-Fi scan with the position $(x, y, z)$ in the built map.
In the \textit{online phase}, the re-localization framework, running on-board the robotic platform, uses a single LiDAR and Wi-Fi scan as the input data to localize itself on the built map and to start the new mission.
The proposed solution has been developed with aiming at minimizing the weight and the space required by the overall system, thus allowing the integration on several possible robotics platforms, spacing from rovers (e.g., Pioneer 3-AT), legged robots (e.g., Unitree Go1) and even compact UAVs (e.g., Holybro X500 V2), as depicted in Fig.~\ref{fig:introduction}. The developed system architecture can be divided into two main hardware components, the IEEE~$802.11$~$2.4$~GHz Wi-Fi scanner and the 3D LiDAR scanner responsible for the localization through a SLAM algorithm, both detailed in the following.
\subsection{$2.4$~GHz Wi-Fi Scanner and 3D LiDAR} \label{subsec:scanner_lidar}
\vspace{-0.25mm}
In order to collect all the information regarding the Wi-Fi networks used to build the fingerprints' database, a suitable compact Wi-Fi scanner has been developed. Given the need to minimize the device footprint, on the hardware side, as Wi-Fi scanner we decided to use a compact-size development board based on the ESP32 SoC, which is able to provide both IEEE~$802.11$ Wi-Fi and Bluetooth connectivity on the $2.4$~GHz RF band. Despite Wi-Fi connectivity built-in on the used robotics platforms, we decided to rely on an external additional adapter in order to (i) have a detachable solution deployable on the interested platform, (ii) avoid impacting the built-in Wi-Fi connectivity, generally used to stream the gathered data, and (iii) minimize the measurement discrepancy introduced by the use of different antennas and Wi-Fi modems, thus allowing to collect cleaner and more reusable data for several experiments. 
The use of the ESP32 programmable micro-controller allowed us to fine tune the Wi-Fi networks scanning process. In fact, a complete scan of all the $2.4$~GHz Wi-Fi channels---$13$ in Europe---requires a relevant amount of time to complete. The Wi-Fi networks scanning process consist in two types of scan: (i) \textit{active scan}, where the scanner sends a \textit{probe request frame} and the nearby APs reply with a \textit{probe response frame}; and (ii) \textit{passive scan}, where the scanner passively listens for incoming \textit{beacon management frames} on each Wi-Fi channel, typically sent by APs every $100$~ms. 
Using the default ESP32 parameters, a Wi-Fi scan of all the~$2.4$~GHz channels requires approximately $2040$~ms, achieving an update frequency of the Wi-Fi data approximately equal to $0.5$~Hz (on average). 
In order to reduce the Wi-Fi scan time and thus increase the update frequency, allowing to collect more data while moving the various robotic platforms and also to make the \textit{offline phase} less tedious and faster, in the firmware developed for the ESP32 board, the Wi-Fi \textit{active scan} time for the channels $1$-$11$ has been reduced from the initial $120$~ms to $85$~ms, while the \textit{passive scan} time for channels $12$ and $13$ has been reduced from $360$~ms to $255$~ms, thus reducing the average complete scan time to approximately~$1445$~ms and consequentially slightly increasing the update frequency of the Wi-Fi data to approximately~$0.69$~Hz (on average). Moreover, to collect more data, the detection of hidden SSID is enabled, allowing to detect both RSSIs and MAC addresses of the hidden networks. All the gathered data (namely: SSID, MAC address, RSSI and channel of each detected AP) from the Wi-Fi scanner are properly sent to the robotic platform companion computer through the USB serial port and, then, published to the custom ROS topic.

For the \textit{first} experiment, the chosen 3D LiDAR scanner is the Velodyne Puck Hi-Res, running at $10$~Hz frequency and featuring $16$ channels with a $20^{\circ}$ vertical Field Of View (FOV) and a $360^{\circ}$ horizontal FOV. In the \textit{second} experiment, we use the Ouster OS1-32 LiDAR, running at $10$~Hz frequency and featuring $32$ channels with a $45^{\circ}$ vertical FOV and a $360^{\circ}$ horizontal FOV. Along with the integrated IMU unit, they are used to run DLO~\cite{DLO}, a SLAM algorithm responsible for providing the odometry. For the 3D LiDAR-based place recognition, discussed in Section~\ref{sec:methodology}, we train a model similar to OREOS~\cite{oreosSchaupp2019}, but with a different backbone model. In particular, instead of using a $3$~layer CNN, we use ResNet18 to get a deeper and more sophisticated feature extraction, assuring a robust performance even with sparse LiDAR scans. The model was trained using data collected from multiple runs in the underground corridors of Lule\r{a} University of Technology (LTU) and a real-life underground mine facility located in Lule\r{a}, Sweden. 
\subsection{Datasets and Platforms} \label{subsec:dataset_platforms}
The experimental evaluations took place in two different environments with two different platforms. In the \textit{first} scenario, we test in the underground corridors of LTU with a Spot robot manufactured by Boston Dynamics, and equipped with the aforementioned sensors, featuring an Intel NUC on-board computer with an Intel Core i5-10210U and 8~GB of RAM. This urban environment offers long, self-similar corridors with some parts containing glass, doors and various objects. Being part of the university, the corridors are populated with fixed Wi-Fi APs as well as temporary APs from various devices. During this experiment, the robot traversed approximately $500$~m at a constant speed of $5$~km/h.

The \textit{second} experiment took place in a modern underground mining facility, being fully equipped with a Wi-Fi infrastructure composed of several APs. The same sensors have been mounted on top of a vehicle, which then navigated through the mine at various speeds (between $15$-$20$~km/h) for a total distance of approximately equal to~$1$~km. In both these experiments, we test the re-localization performance continuously as the robotic platform navigates the environment. The sampling speed of the LiDAR and the Wi-Fi scanners is determined by the slowest one, which in both cases is the Wi-Fi scanner. Therefore, the results presented below are from a continuous evaluation for every sampling step defined by the rate of the Wi-Fi scanner.
\subsection{Results and discussion}
\begin{figure*}[t!]
    \centering
    \includegraphics[width=\textwidth]{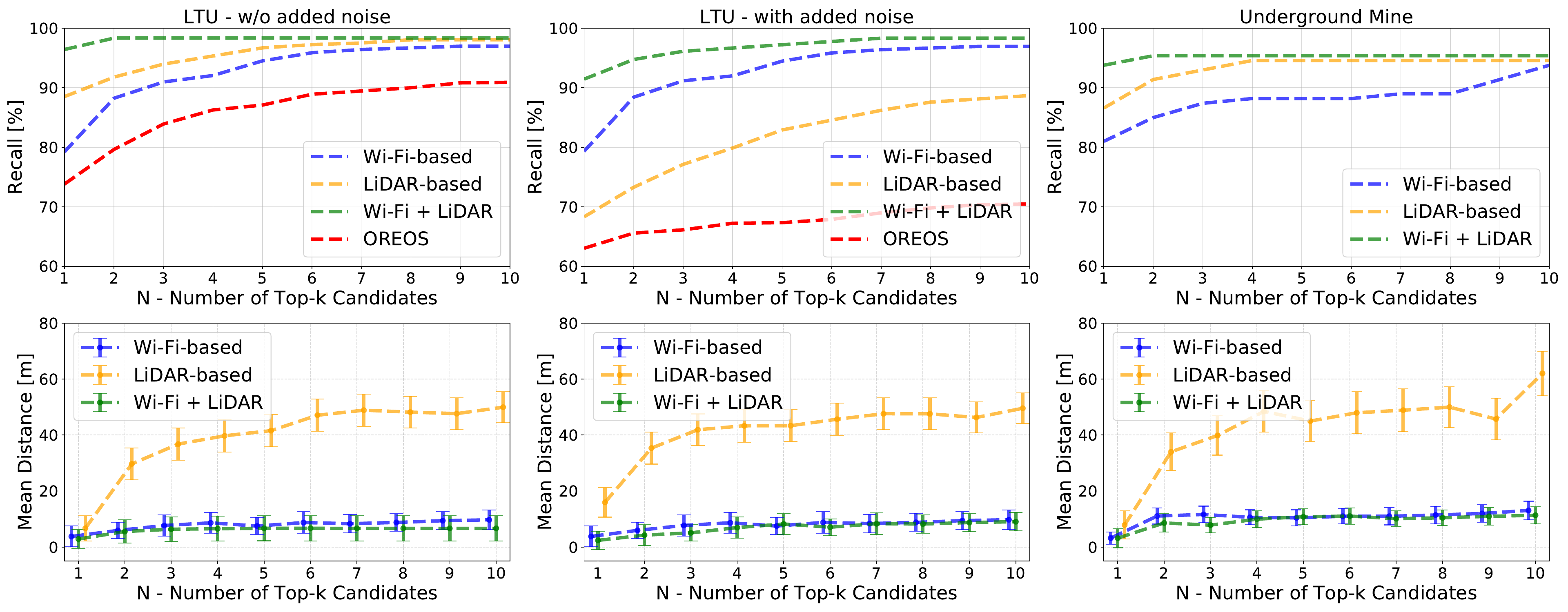}
    \caption{Top plots showcase the results from the real-world experiments in terms of place recognition for an increasing number $N$ of nearest place candidates retrieved from the map. A candidate is considered correct if it is within a $3$~m radius from the corresponding pose in the database. Bottom plots demonstrate the mean distance of the predicted pose to the ground truth from the map, for each candidate, with the addition of the min and max deviations.}
    \label{fig:recall_mean_std}
    \vspace{-5mm}
\end{figure*}

\begin{figure}[h!]
    \centering
    \includegraphics[width=\columnwidth]{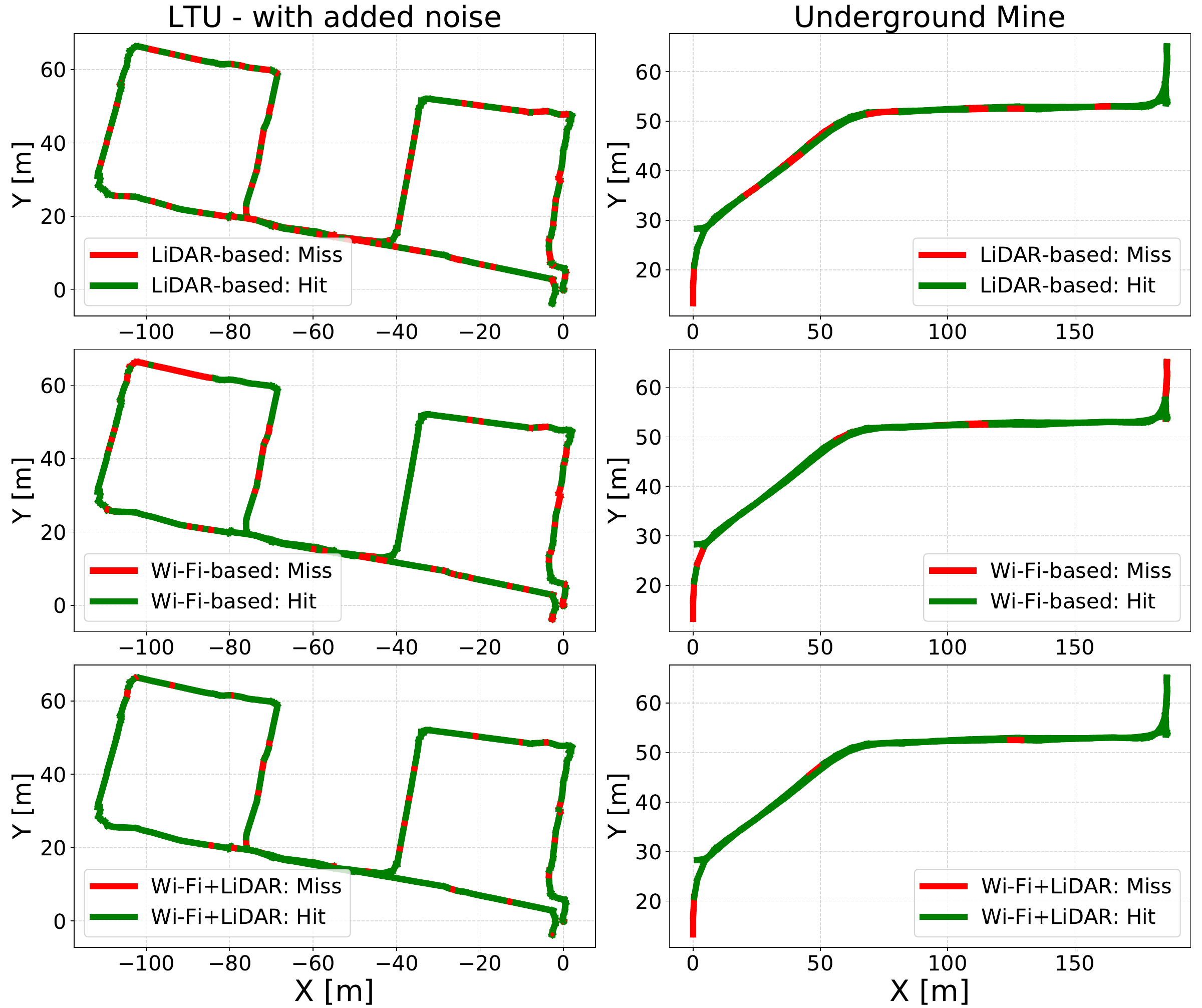}
    \caption{Trajectory for each experiment, based on the \textit{top-1} candidate (green points represent the correct predictions, red points denote the incorrect ones).}
    \label{fig:trajectories}
    \vspace{-5mm}
\end{figure}

In this section, we go through the results that are presented in Fig.~\ref{fig:recall_mean_std}, Fig.~\ref{fig:trajectories} and Table~\ref{table:recall_mean_std}. The metrics include (i) the recall score for the increasing number of nearest place candidates retrieved from the map and (ii) the mean and standard deviation of the distance from the current pose to the one queried from the database. The latter metrics heavily relies on the recall score and on the density of the sampling space. 
Starting from the results with the \textit{first} experiment at the urban indoor environment, shown in Fig.~\ref{fig:recall_mean_std} and Table~\ref{table:recall_mean_std}, it can be observed how the LiDAR based method demonstrates a good performance in terms of recall percentage. On the other hand, the mean distance of the predicted pose, compared to the ground truth within the map, is significantly high, especially for the increasing number of \textit{top-k} candidates. On the contrary, for the Wi-Fi fingerprinting, the mean distance is maintained low (around $3$~m, increasing to approximately $5$-$10$~m as expected for the less likely candidates). The proposed switching solution is then able to combine the best achieved from both approaches (namely, the good recall percentage of the LiDAR-based method and the lower mean distance error of the Wi-Fi fingerprinting), finally resulting in a $97\%$ recall and a mean error of $3.3$~m for the best candidate selection. This performance gain is explained by the nature of the Wi-Fi fingerprinting, being able to narrow down the search space, compared to LiDAR, that is prune to perceptual aliases. 
In 3D point clouds, descriptors are used to capture and represent the unique features of each point. However, it is possible for two descriptors to be similar despite originating from different locations if the underlying features of the point cloud, such as shape or texture, exhibit similarities. This similarity can occur because the descriptors are designed to extract and encode relevant information about the point cloud, allowing them to capture similar patterns or characteristics regardless of their spatial origin.
The deterministic nature of the Wi-Fi fingerprinting overcomes this issue by removing the outlier candidates, and the combination of both yields the best result.

To further evaluate the performance of our novel framework, we degrade the performance of the LiDAR by adding Gaussian Noise at $40\%$ of the samples with $\sigma_n^2$ = $0.015$. In the second column of Fig.~\ref{fig:recall_mean_std}, the performance drop is approximately $20\%$ for the recall percentage and the mean error distance from $7$~m to $16$~m. At the opposite, with the combined method the decrease is only $6\%$, demonstrating the ability to decide between the best candidates in the presence of disturbance. For both scenarios, in the urban environment, the performance of OREOS showcases the need for a deeper feature extraction pipeline, leveraging its proposed point cloud to range image projection. For the rest of the experiments, we disregard this method.

The \textit{second} experiment took place in a real-world underground mining facility that contains long featureless tunnels with multiple drifts. In the last column of Fig.~\ref{fig:recall_mean_std}, the results of the switching solution further validate our hypothesis. In detail, the addition of the Wi-Fi can narrow down the search space, and the resulting common candidates yield a $93\%$ recall score a mean error distance approximately equal to $3.4$~m. The high recall score for the LiDAR-based method is due to the higher resolution of the LiDAR scanner, which, in this case, was the OS1-32 and had a number of channels double with regard to the VLP16. Finally, as shown Fig.~\ref{fig:trajectories}, the red mismatched places are corrected, as long as at least one of the components predicted it right. 


\section{FUTURE DEVELOPMENTS} \label{sec:improvements}
There are several ways to improve the accuracy of the proposed framework. For example, in order to increase the amount of Wi-Fi data collected during the \textit{offline} phase of the fingerprinting process, multiple scanners based on the same development board could be used, therefore increasing the features available for the framework. Moreover, the newer Wi-Fi standards, namely IEEE~$802.11$ac (Wi-Fi~$5$) and IEEE~$802.11$ax (Wi-Fi~$6$E), introduce the use of the additional $5$~GHz and $6$~GHz RF bands alongside the $2.4$~GHz band, with the aim to improve the throughput of wireless networks. An enhanced version of the framework could adopt a newer Wi-Fi sensor able to scan multiple bands, especially the widely used $5$~GHz band, that, given the shorter range, should return more \textit{granular} fingerprints, further improving the $2.4$~GHz solution. Besides Wi-Fi networks, 5G cellular networks adoption is rapidly increasing, with the continuous deployment of indoor private 5G networks based on small Base Transceiver Stations (BTSs) for industrial applications in various environments. Given the different radio technologies used by 5G, the integration of the cellular network fingerprint can further enhance the accuracy of the framework. Finally, more advanced ML techniques can be adopted\cite{mlwifi2021, deeplearning2021}, allowing a better usage of the gathered data to further increase the accuracy.









\section{CONCLUSIONS}
\label{sec:conclusions}


This paper has proposed a framework that integrates LiDAR-based descriptors and Wi-Fi fingerprinting to overcome the global localization challenge in pre-mapped environments. The preliminary experimental evaluations, as shown in Section~\ref{sec:experimental}, are promising. These evaluations highlight the framework's potential, particularly for localization applications in environments where pure LiDAR solutions may encounter technological limitations. By exploiting the strengths of both LiDAR SLAM and Wi-Fi fingerprinting, the framework consistently achieves improved performance, especially in challenging environments where traditional methods fall short. Experimental results from urban indoor and underground mining scenarios validate the effectiveness of the switching solution, which combines the LiDAR-based method's high recall percentage with the Wi-Fi fingerprinting's lower mean distance error. The proposed framework holds promise for practical implementation in inspection and monitoring scenarios, empowering mobile robots with accurate global localization capabilities.


\addtolength{\textheight}{-1cm}   


\bibliographystyle{./IEEEtranBST/IEEEtran}
\bibliography{./IEEEtranBST/IEEEabrv,references}

\note{
}

\end{document}